\def\year{2021}\relax
\documentclass[letterpaper]{article} % DO NOT CHANGE THIS
\usepackage{aaai21_arXiv}  % DO NOT CHANGE THIS
\usepackage{times}  % DO NOT CHANGE THIS
\usepackage{helvet} % DO NOT CHANGE THIS
\usepackage{courier}  % DO NOT CHANGE THIS
\usepackage[hyphens]{url}  % DO NOT CHANGE THIS
\usepackage{graphicx} % DO NOT CHANGE THIS
\usepackage{graphicx}  % DO NOT CHANGE THIS
\usepackage{natbib}  % DO NOT CHANGE THIS AND DO NOT ADD ANY OPTIONS TO IT
\usepackage{caption} % DO NOT CHANGE THIS AND DO NOT ADD ANY OPTIONS TO IT
\usepackage[switch]{lineno}  %
\title{\textsc{TripleTree}: A Versatile Interpretable Representation\\of Black Box Agents and their Environments}
\author{Tom Bewley\thanks{Supported by an EPSRC/Thales industrial CASE award.} and Jonathan Lawry}
\affiliations{
    Department of Engineering Mathematics\\
    University of Bristol\\
    Bristol, United Kingdom\\
    \{tom.bewley,\ j.lawry\}@bristol.ac.uk

}
\usepackage[colorlinks]{hyperref}
\usepackage{xcolor}
\usepackage{pdfpages}
\definecolor{linkcolour}{HTML}{B24A17}%{16585A}
\definecolor{citecolour}{HTML}{B24A17}%{16585A}
\definecolor{urlcolour}{HTML}{B24A17}%{16585A}
\hypersetup{linkcolor=linkcolour,citecolor=citecolour,filecolor=urlcolour,urlcolor=urlcolour}
\usepackage{amsmath,amssymb,amsfonts}

\DeclareMathOperator*{\argmax}{argmax}
\begin{document}

\maketitle

\begin{abstract}
In explainable artificial intelligence, there is increasing interest in understanding the behaviour of autonomous agents to build trust and validate performance. Modern agent architectures, such as those trained by deep reinforcement learning, are currently so lacking in interpretable structure as to effectively be black boxes, but insights may still be gained from an external, behaviourist perspective. Inspired by conceptual spaces theory, we suggest that a versatile first step towards general understanding is to discretise the state space into convex regions, jointly capturing similarities over the agent's action, value function and temporal dynamics within a dataset of observations. We create such a representation using a novel variant of the CART decision tree algorithm, and demonstrate how it facilitates practical understanding of black box agents through prediction, visualisation and rule-based explanation.

\end{abstract}

\section{Introduction}

This paper explores representational tools for understanding the behaviour of extant autonomous agents while treating them and their environments as black boxes. In popular taxonomies of explainable artificial intelligence (XAI), this is categorised as post hoc, model-agnostic, global explanation \cite{adadi2018peeking}. While black box behaviourist analysis is inherently limited  \cite{chomsky1959review}, it can nonetheless serve the practical goals of XAI, which include building trust among human stakeholders and validating the performance of safety-critical systems. It is also viable in contexts where theoretical understanding of the agent's internal mechanism is lacking, as with modern deep learning systems \cite{samek2017explainable}, or where access to this mechanism is impractical or restricted. 

We introduce a new data-driven model of black box agents, called \textsc{TripleTree}, which builds on the flexible and interpretable architecture of a binary decision tree. As such, it provides a powerful tool for answering many meaningful questions about agent behaviour.

\subsection{Problem Setup}

Suppose that we need to understand the behaviour, performance and possible failure modes of an autonomous agent operating within a dynamic and complex environment. A priori, we know nothing of the agent's provenance -- its governing policy may be a product of reinforcement learning (RL), optimal control algorithms, evolution, or explicit manual design -- but we assume that it can be analysed using the theoretical formalism of a \textit{Markov decision process} (MDP). We refer readers to \cite{sutton2018reinforcement} for an overview of MDPs, and adopt the MDPNv1 notational standard \cite{thomas2015notation}. 

To learn anything about the agent we must gather data, and in doing so we make the \textit{black box assumption} \cite{wachter2017counterfactual,guidotti2019factual,coppens2019distilling}. That is, we take the role of an observer of the agent-environment complex, with no access to the internal structure of either system, but the ability to record environment states $s\in\mathcal{S}$, agent actions $a\in\mathcal{A}$, and instantaneous rewards $r\in\mathcal{R}$, and their order of occurrence. We thereby assemble $\mathcal{D}$, an ordered dataset of triplets $(S_t,A_t,R_t)$. $S_t$, $A_t$ and $R_t$ are the state, action and reward from one timestep, uniquely indexed by $t$. If the MDP is episodic (see \cite{sutton2018reinforcement}), we keep a record of which states are the initial and terminal ones in each episode. Importantly, we assume that states are represented by vectors of real-valued features, each with a straightforward semantic interpretation (such as a physical quantity). In doing so, we bypass a challenging phase of state representation learning.\footnote{ We address the problem of interpretable feature construction and selection in \cite{bewley2020modelling}.}

The data in $\mathcal{D}$ are generated by the interaction of opaque and complex mechanisms. How might they get us to a position of understanding? Following prior work across the academic spectrum \cite{carnap1967logical,rosch1976basic,edelman1998representation}, we take the view that understanding arises by searching for similarities in observed data. A variant of this idea is G\"{a}rdenfors' theory of \textit{conceptual spaces} \cite{gardenfors2004conceptual}.

\subsection{Conceptual Spaces and Decision Trees}

G\"{a}rdenfors views sensory observations as embedded in high-dimensional mathematical spaces, and proposes that the building blocks of abstract reasoning are convex regions of such spaces, within which all contained observations are similar according to some salient measure. Such regions are deemed \textit{natural properties} of the system being observed, and can be combined to form semantic concepts such as objects, categories, actions and events. For our purposes, the observations $\mathcal{D}$ do indeed lie within a mathematical space: the MDP state space $\mathcal{S}$. We consider how these observations can be grouped into convex regions of $\mathcal{S}$ based on context-specific measures of similarity. Rather than talking about the system on a state-by-state basis, we may then analyse such regions as meaningful entities in themselves, within which the agent behaves in predictable ways, and between which it moves in predictable patterns. This is a kind of MDP \textit{state abstraction}, informed by observations of the agent itself.

A central issue is choosing which attributes to use for measuring similarity within regions. An obvious criterion is the agent's action. If the agent takes the same action throughout a significant region of $\mathcal{S}$, then it seems that region is worthy of being explicitly represented. Alternatively, we might be interested in the agent's performance as measured by the reward function, and thus measure similarity using the reward elements in $\mathcal{D}$. In practice, it is likely more informative to invoke the notion of value, which is the expected sum of reward after the agent visits each state, temporally discounted by $\gamma\in[0,1]$. An empirical value estimate can be computed for each sample $t\in\mathcal{D}$ using the rewards of successive samples: $V_t=\sum_{k=0}^T\gamma^kR_{t+k}$. Here, $T$ is the time until termination in an episodic MDP, and $\infty$ otherwise. A third valid option is to define similarity via the temporal dynamics of the MDP itself. These can be neatly captured by the time derivatives of state features, which in discrete time systems are equal to the change in state between timesteps. For each sample $t\in\mathcal{D}$, we define this as $S_{t+1}-S_t$.

Our key assertion in this paper is that there is no need to choose between these three sources of similarity, and that a powerful and versatile model results from identifying regions of $\mathcal{S}$ that are similar from \textit{all three} perspectives. We complete our model by calculating transition probabilities between regions -- the probability that having being observed at a state in one region, the agent will move to another region next. These effectively define a probabilistic finite state machine (FSM) model of the agent-environment dynamics, and can be estimated by harnessing the temporal ordering of $\mathcal{D}$.

\begin{figure}[h!]
\includegraphics[width=\columnwidth]{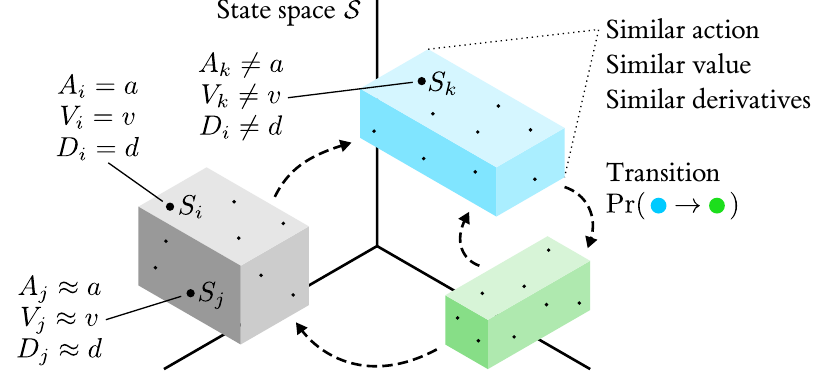}
\caption{A general model of agent-environment dynamics.}
\label{fig:diagram}
\end{figure}

Figure \ref{fig:diagram} summarises the proposed model. As we hope to demonstrate, it enables us to make sense of pertinent questions about the key invariances and changes points in the agent's behaviour, the environmental factors responsible for this behaviour and perturbations which would alter it, the regions of $\mathcal{S}$ most commonly visited, and the most likely trajectories between particular states of interest. 

\pagebreak

We now turn to the question of practical implementation. There exists an ideal computational tool for finding internally-similar convex regions of mathematical spaces: the humble \textit{decision tree}. A decision tree is `grown' to predict an output label by recursive binary partitioning of a space of input features. Typically, partitions are axis-aligned, so each resultant region (corresponding to a leaf of the tree) is a hyperrectangle. Partitions are chosen to greedily minimise a measure of impurity in the output labels of training data at each node, which is tantamount to finding and preserving similarity as conceived in our conceptual space model. Trees are popular in XAI, and often hailed as the gold standard of interpretability \cite{wan2020nbdt}. Their hierarchical structure means that global functionality can be analysed as the concatenation of local effects, without higher-order interaction. Their alternative representation as rule sets in disjunctive normal form also enables factual and counterfactual explanation of their outputs \cite{guidotti2019factual}. 

The sub-field of agent explainability has seen enthusiastic uptake of decision trees  \cite{bastani2018verifiable,coppens2019distilling,bewley2020modelling}. However, these works focus exclusively on predicting and explaining a black box agent's single next action in any given state, effectively approximating its policy function, denoted by $\pi$. This approach uses only the first two elements of the triplets $(S_t,A_t,R_t)$ in $\mathcal{D}$, and ignores their ordering, thus foregoes an opportunity for far richer analysis. The all-important notion of value, as well as the temporal dynamics of agent-environment interaction, are entirely absent from a policy-only model, which in isolation is insufficient for answering many reasonable questions about the target agent's performance and dynamical properties. 

This lack of versatility is a product of how data are stored in the tree, and of the algorithm used to grow it, rather than an inherent limitation of the decision tree paradigm. We propose to extend the standard notion of impurity to capture multiple facets agent-environment interaction, making full use of the data in $\mathcal{D}$, and providing a practical implementation of the conceptual model in figure \ref{fig:diagram}. Concretely, we grow the tree using a hybrid of three impurity measures related to the agent's action, expected value and state feature time derivatives, hence our new model's name: \textsc{TripleTree}. By modifying a weight vector $\theta\in\mathbb{R}_+^3$, which sets the influence of the three measures, we can smoothly trade off between three types of interpretable model of the system: 
\begin{itemize}
\item $\theta=[1,0,0]$: A conventional policy-only model.
\item $\theta=[0,1,0]$: A value function.
\item $\theta=[0,0,1]$: A model of the environment state dynamics.
\end{itemize}
Any other weighting gives a blended combination of the three, allowing for multifactorial analysis.

\section{The \textsc{TripleTree} Model}

\subsection{Basic Structure: CART}

\textsc{TripleTree} is an extension of the CART algorithm \cite{CART}, which we briefly introduce first. We assume that given a dataset $\mathcal{D}$, CART is being used to predict the agent's action given the state; a policy-only model.

Let $\mathcal{I}=\{1,...,|\mathcal{D}|\}$ be the set of timestep indices in $\mathcal{D}$, used to initialise the tree's root node before any partitions are made. Let $\mathcal{I}_N$ be the subset of $\mathcal{I}$ at any given node $N$. To split this node in two, thereby growing the tree, CART searches over binary partitions $\mathcal{I}_{N}=\{\mathcal{I}_0,\mathcal{I}_1\}$ such that for some state feature $f$ and numerical threshold $\tau\in\mathbb{R}$:
\begin{equation}
(\forall t\in \mathcal{I}_0\ S_t^{(f)}<\tau)\ \land\ (\forall t'\in \mathcal{I}_1\ S_{t'}^{(f)}\geq\tau).
\end{equation}
Here, $S_t^{(f)}$ is the $f$th element of the state vector $S_t$. For each candidate partition, we calculate the population-weighted reduction in a measure of action label impurity $\text{I}$ induced by dividing the set into these two parts. For a discrete action space $\mathcal{A}$, the Gini impurity is used:
\begin{equation} \label{eq:gini_act}
\text{I}_N=\frac{1}{|\mathcal{I}_N|^2}\sum_{a\in\mathcal{A}}\text{count}(\mathcal{I}_N,a)(1-\text{count}(\mathcal{I}_N,a)),
\end{equation}
where $\text{count}(\mathcal{I}_N,a)=|\{t\in\mathcal{I}_N:A_t=a\}|$. For continuous actions $\mathcal{A}=\mathbb{R}$, the impurity measure is the variance:
\begin{equation} \label{eq:var_act}
\text{I}_N=\frac{1}{2|\mathcal{I}_N|^2}\sum_{t\in\mathcal{I}_N}\sum_{t'\in\mathcal{I}_N}(A_t-A_{t'})^2.
\end{equation}
The quality $\text{Q}$ of a candidate partition is defined as:
\begin{equation} \label{eq:part_quality}
\text{Q}(\mathcal{I}_N,\{\mathcal{I}_0,\mathcal{I}_1\})=\text{I}_{N}-\frac{\text{I}_0|\mathcal{I}_0|+\text{I}_1|\mathcal{I}_1|}{|\mathcal{I}_{N}|}.
\end{equation}
CART selects the partition that maximises $\text{Q}$, and the corresponding feature $f$ and threshold $\tau$ are recorded in the tree. $\mathcal{I}_0$ and $\mathcal{I}_1$ also define the members of two new child nodes of $N$. CART proceeds to search for the best binary partition of each child, and grows the tree depth-first up to a stopping condition, such as a depth limit. 

In any tree, a subset of nodes, called the \textit{leaves} $\mathcal{L}$, remain childless. Every sample in $\mathcal{D}$ is a member of exactly one leaf $L$, whose key attribute is an action prediction $\tilde{a}_L\in\mathcal{A}$. For discrete $\mathcal{A}$, this is typically the modal action among the leaf's constituent samples: $\tilde{a}_L=\operatorname{argmax}_a\text{count}(\mathcal{I}_L,a)$. For continuous $\mathcal{A}$, the mean is used: $\tilde{a}_L=\sum_{t\in\mathcal{I}_L}A_t/|\mathcal{I}_L|$. To predict an action for an unseen state vector $s$, the tree propagates $s$ down a path from the root by comparing its features to the thresholds encountered, then returns the prediction of the leaf which is ultimately reached. 

\subsection{Modifications and Extensions}

Our fundamental divergence from CART is in the use of the entire content and strucure of $\mathcal{D}$. \textsc{TripleTree} accepts ordered triplet samples of the form $(S_t,A_t,R_t)$, and prior to commencing growth evaluates two additional attributes, namely the value estimate $V_t=\sum_{k=0}^T\gamma^kR_{t+k}$ and state derivative vector $D_t=S_{t+1}-S_t$. Rather than using only the agent's action as an output label, each leaf $L$ is associated with three predictions: the action $\tilde{a}_L$, a value estimate $\tilde{v}_L$ (the mean of the leaf's contituent samples), and a state derivative estimate $\tilde{d}_L$ (the elementwise mean).

The \textsc{TripleTree} growth algorithm trades off the ability to make these three kinds of prediction by encouraging leaves to have low variability in all three attributes across their constituent samples. To achieve this, we compute three measures of the quality of candidate partitions:
\begin{itemize}
\item \textbf{Action quality} $\text{Q}^{(A)}$: defined exactly as in equation \ref{eq:part_quality}.
\item \textbf{Value quality} $\text{Q}^{(V)}$: defined equivalently, but using the variance in value estimates as the impurity measure $\text{I}^{(V)}$.
\item \textbf{Derivative quality} $\text{Q}^{(D)}$: defined equivalently, but using an impurity measure $\text{I}^{(D)}$ that sums the variance in derivatives across all $d$ of the feature dimensions:
\begin{equation}
\text{I}^{(D)}_N=\frac{1}{2|\mathcal{I}_N|^2}\sum_{f=1}^d\frac{1}{\sigma^{(f)}}\sum_{t\in\mathcal{I}_N}\sum_{t'\in\mathcal{I}_N}(D^{(f)}_t-D^{(f)}_{t'})^2.
\end{equation}
$1/\sigma^{(f)}$ is a normalisation factor for each derivative -- the reciprocal of its standard deviation across $\mathcal{D}$ -- which prevents features with large magnitudes dominating the impurity calculation. This sum-of-variances impurity measure is similar to those used in prior work on multivariate regression trees \cite{de2002multivariate, kim2015decision}.
\end{itemize}

After computing $\text{Q}^{(A)}$, $\text{Q}^{(V)}$ and $\text{Q}^{(D)}$, we aggregate them into a hybrid measure of partition quality $\text{Q}^*$. Having experimented with alternative methods in various MDP contexts, we find that a linear combination provides a good compromise of simplicity, robustness and flexibility:
\begin{equation} \label{eq:part_quality_hybrid}
\text{Q}^*=\left[\frac{\text{Q}^{(A)}_N}{\text{I}^{(A)}_{\text{root}}},\ \frac{\text{Q}^{(V)}_N}{\text{I}^{(V)}_{\text{root}}},\ \frac{\text{Q}^{(D)}_N}{\text{I}^{(D)}_{\text{root}}}\right]\cdot\theta.
\end{equation}
Here we have omitted the arguments of the quality terms for brevity. 
$\theta\in\mathbb{R}_+^3$ is a weight vector, which trades off accurate modelling of the policy, value function and derivatives. Each quality term is normalised by the respective impurity at the root node (i.e. before any partitions are made). This brings the three measures onto equivalent scales. 

CART follows a depth-first growth strategy, which is known to lead to suboptimal allocation of partitions. We depart from this by adopting a simple best-first strategy for selecting which leaf node to partition at each stage of tree growth. Again taking a hybrid view of impurity, we identify the best current leaf to partition, $L_\text{best}$, as follows:
\begin{equation}\label{eq:next_best}
L_\text{best}=\underset{L\in \mathcal{L}}{\argmax}\ |\mathcal{I}_L|\left[\frac{\text{I}^{(A)}_{L}}{\text{I}^{(A)}_{\text{root}}},\ \frac{\text{I}^{(V)}_L}{\text{I}^{(V)}_{\text{root}}},\ \frac{\text{I}^{(D)}_L}{\text{I}^{(D)}_{\text{root}}}\right]\cdot\theta,
\end{equation}
where $\theta$ is the same as in equation \eqref{eq:part_quality_hybrid}. This approach prioritises the partitioning of leaves with high total impurity, weighted by their sample counts. Our criterion for terminating tree growth is a limit on the number of leaves, $|\mathcal{L}|$.

The final feature of \textsc{TripleTree} is the calculation of leaf-to-leaf transition probabilities. Let $\text{leaf}(t)=L\in\mathcal{L}:t\in\mathcal{I}_L$ be the leaf at which a sample $t$ resides. For terminal samples in episodic MDPs, we define $\text{leaf}(t)=\emptyset$. Furthermore, let $\mathcal{I}^*_L=\{t\in\mathcal{I}_L:\text{leaf}(t-1)\neq L\}$ be the subset of $\mathcal{I}_L$ whose predecessors are not themselves in $\mathcal{I}_L$: the first in each sequence of successive observations that reside at $L$. We perform our calculations at the level of sequences, rather than individual samples, to avoid a double-counting effect. For each sequence-starting sample $t\in\mathcal{I}^*_L$, we find the length of its successor sequence, $\text{seqlen}(t)$, and the leaf containing the sample that breaks it, $\text{nextleaf}(t)$:
\begin{equation}
\begin{array}{c}
\vspace{0.2cm}
\text{seqlen}(t)=\min\{k:\text{leaf}(t+k)\neq\text{leaf}(t)\}\ ;\\
\text{nextleaf}(t)=\text{leaf}(t+\text{seqlen}(t)).
\end{array}
\end{equation}
For a given source leaf $L$ and destination leaf $L'$, we are interested in the subset of $\mathcal{I}^*_L$ whose successor sequences are followed by a transition to $L'$:
\begin{equation}
\mathcal{I}^*_{L\rightarrow L'}=\{t\in\mathcal{I}^*_L:\text{nextleaf}(t)=L'\}.
\end{equation}
Note that in episodic MDPs, $\mathcal{I}^*_{L\rightarrow \emptyset}$ is well-defined and meaningful; it contains the members of $\mathcal{I}^*_{L}$ for whom the episode terminates before a transition to another leaf. We can now compute the empirical probability that any given sequence in $L$ ends in a transition to $L'$, and the mean length of such a sequence:
\begin{equation}
P_L(L')=\frac{|\mathcal{I}^*_{L\rightarrow L'}|}{|\mathcal{I}^*_{L}|}\ ;\ T_L(L')=\sum_{t\in \mathcal{I}^*_{L\rightarrow L'}}\frac{\text{seqlen}(t)}{|\mathcal{I}^*_{L\rightarrow L'}|}.
\end{equation}
Transition probabilities and times are stored at their respective source leaves ($L$ here) as further attributes alongside the predictions $\tilde{a}_L$, $\tilde{v}_L$ and $\tilde{d}_L$.

In summary, the key features of \textsc{TripleTree} are:
\begin{itemize}
\item Acceptance of the triplet observations in $\mathcal{D}$, calculation of value and state derivatives for each sample, and storage of predicted values of these variables at each leaf.
\item A hybrid measure of partition quality $\text{Q}^*$, mediated by a weight vector $\theta$, which trades off the tree's abilities to predict the target agent's action, value and state derivatives.
\item Calculation of $P_L$ and $T_L$ to encode information about temporal dynamics in terms of leaf-to-leaf transitions.
\item A best-first growth strategy.
\end{itemize}

A Python implementation of \textsc{TripleTree} is available on GitHub at \url{https://github.com/tombewley/TripleTree}.

\section{Related Work}

Before the widespread adoption of deep neural network function approximators in RL, decision trees were used to create discretised state abstractions for tabular Q-learning algorithms \cite{uther1998tree}. Tree models have also been used to learn a value function with the aim of creating an interpretable agent that performs well in the task environment \cite{pyeatt2003reinforcement,roth2019conservative}, and also to mimic the value function of an existing black box policy as a route to explainability \cite{liu2018toward}. This latter model also keeps track of transition probabilities between tree leaves, similarly to our approach. In \cite{jiang2019experience}, a decision tree is grown to minimise the impurity of environment state derivatives as part of a model-based RL framework, and in \cite{kim2015decision} a tree is optimised for sequential prediction by jointly minimising loss on consecutive timesteps. Other work has looked at approximating a recurrent neural network policy as a finite state machine for visualisation and analysis \cite{koul2018learning}. 

We know of one work that considers a hybrid action- and value-based tree impurity measure \cite{saghezchi2010multivariate}, but the idea is tangential to the main topic of the paper and its implications left unconsidered. We are unaware of any prior work that jointly represents the policy, value function and temporal dynamics in one decision tree, or considers the benefits of doing so for interpretability. 

\section{Prediction Tradeoff Experiments}

\begin{figure}
\centering
\includegraphics[width=0.8\columnwidth]{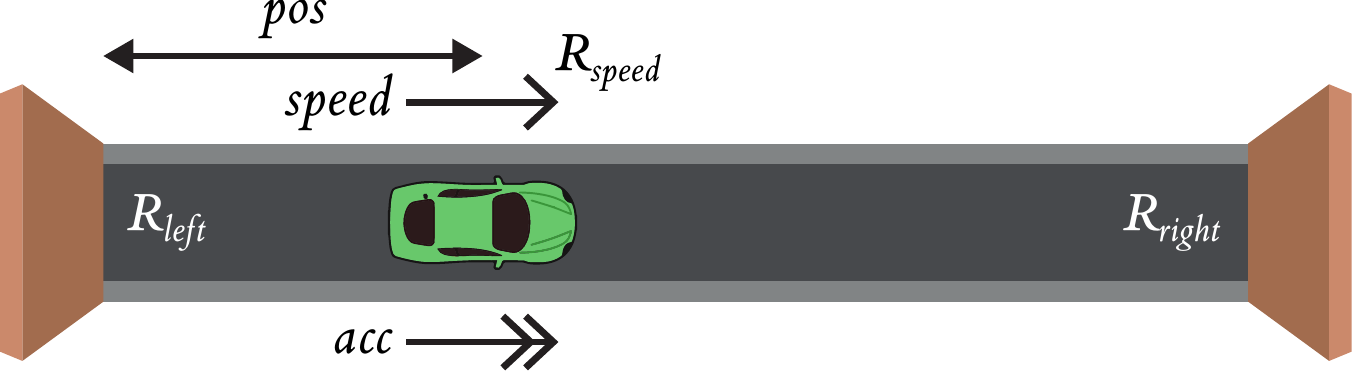}
\caption{The 2-dimensional road MDP.}
\label{fig:road}
\end{figure}

We initially validate \textsc{TripleTree} in a simple MDP with $2$ state features and $2$ discrete actions. This can be interpreted as a straight road, down which a vehicle agent can drive in either direction. The state features are position $pos\in[0,3]$ (increasing left-to-right) and speed $speed\in[-0.1,0.1]$, and the agent's action is a small positive or negative acceleration $acc\in\{-0.001,0.001\}$. Walls lie at the left and right ends of the road; a collision with either yields a reward of $R_{left}$ and $R_{right}$ respectively and instantly terminates the simulation episode. The agent also receives reward in each non-terminal state in proportion to its absolute speed: $R_{speed}\times|speed|$. Figure \ref{fig:road} summarises this information. 

For a given $R_{left}$, $R_{right}$, $R_{speed}$, discount factor $\gamma$ (we use $\gamma=0.99$), and suitable discretisation of $\mathcal{S}$ (we use a $30\times 30$ grid) an optimal policy can be found by dynamic programming (DP). We use the DP policies for four reward function variants as the target agents in our experiments. For each, we create a dataset $\mathcal{D}$ with $10^4$ samples, by running randomly-initialised episodes of $100$ timesteps.

\begin{figure}
\includegraphics[width=\columnwidth]{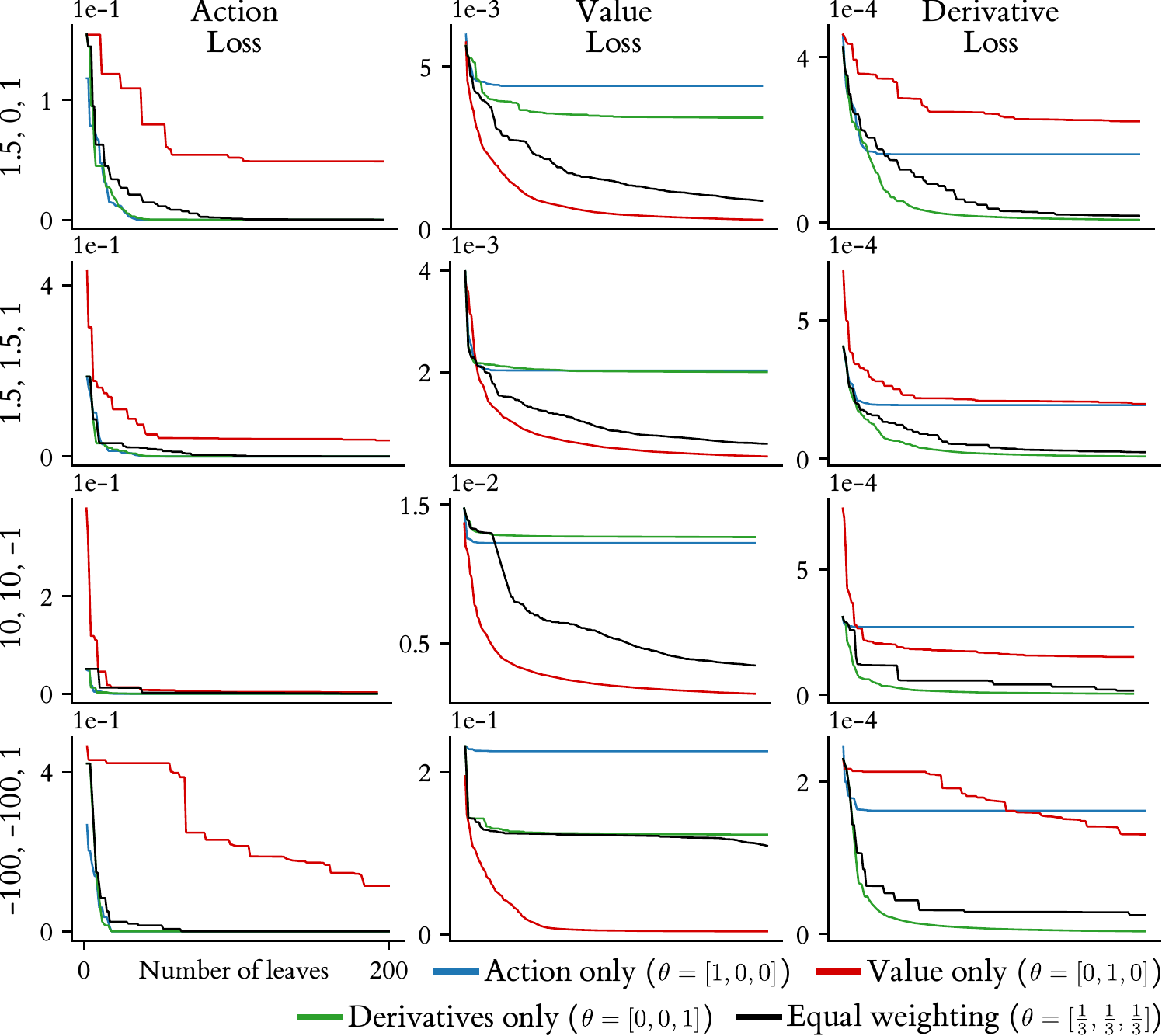}
\caption{Prediction losses for four variants of the road MDP, with $R_{left}$, $R_{right}$, $R_{speed}$ as stated in the left-hand labels.}
\label{fig:loss_curves}
\end{figure}

Figure \ref{fig:loss_curves} shows the result of growing a \textsc{TripleTree} of up to $200$ leaves using these four datasets, with  
various impurity weightings $\theta$. The columns show predictive losses for action (proportion of incorrect predictions), value (RMS error) and derivatives (dot product of RMS error with normalisation factors $1/\sigma$) as a function of leaf count. Naturally, different trees result when different $\theta$ vectors are used, and in all cases the lowest loss of each type is obtained by exclusively using the corresponding partition quality measure. Crucially, however, using an equal weighting ($\theta=[1/3,1/3,1/3]$; black curves) offers a strong compromise between the three modes of prediction. For action and derivatives, equal weighting converges slower than exclusive weighting, but to virtually the same asymptotic loss, with the greatest disparity for trees with around $50$-$100$ leaves. For value the gap is more significant. This indicates that in this MDP, there tend to be regions of $\mathcal{S}$ in which value varies significantly but the agent's action and state derivatives do not, thereby creating a conflict as to which leaves are worthy of partitioning. This phenomenon is most pronounced for the policy on the bottom row. Another notable trend is that partitioning on derivatives alone does very well in terms of action loss. This makes perfect sense once we realise that the agent's action ($acc$) is exactly the time derivative of one of the state features ($speed$).

This analysis begs the question: what is the best $\theta$ for this MDP? Ultimately, the answer depends on the intended application, but if general versatility is important then we may wish to minimise the \textit{worst} of the three loss types. Our analysis along these lines (see Appendix A) suggests that in this MDP, a good compromise is attained by placing increased weight on value impurity: $\theta=[0.2,0.6,0.2]$.

\section{Multiattribute Visualisation in State Space}

Recall that in a decision tree, each leaf is associated with a hyperrectangle in the $d$-dimensional state space $\mathcal{S}$, whose boundaries correspond to the partitions of its ancestor nodes. If $d\in\{1,2\}$, hyperrectangles reduce to lines or rectangles, which can be directly shown on axes corresponding to $\mathcal{S}$ itself (we return to the $d>2$ case later in this paper). Each leaf can be coloured according to some salient attribute including, but not limited to, one of its three predictions.

\begin{figure}
\includegraphics[width=\columnwidth]{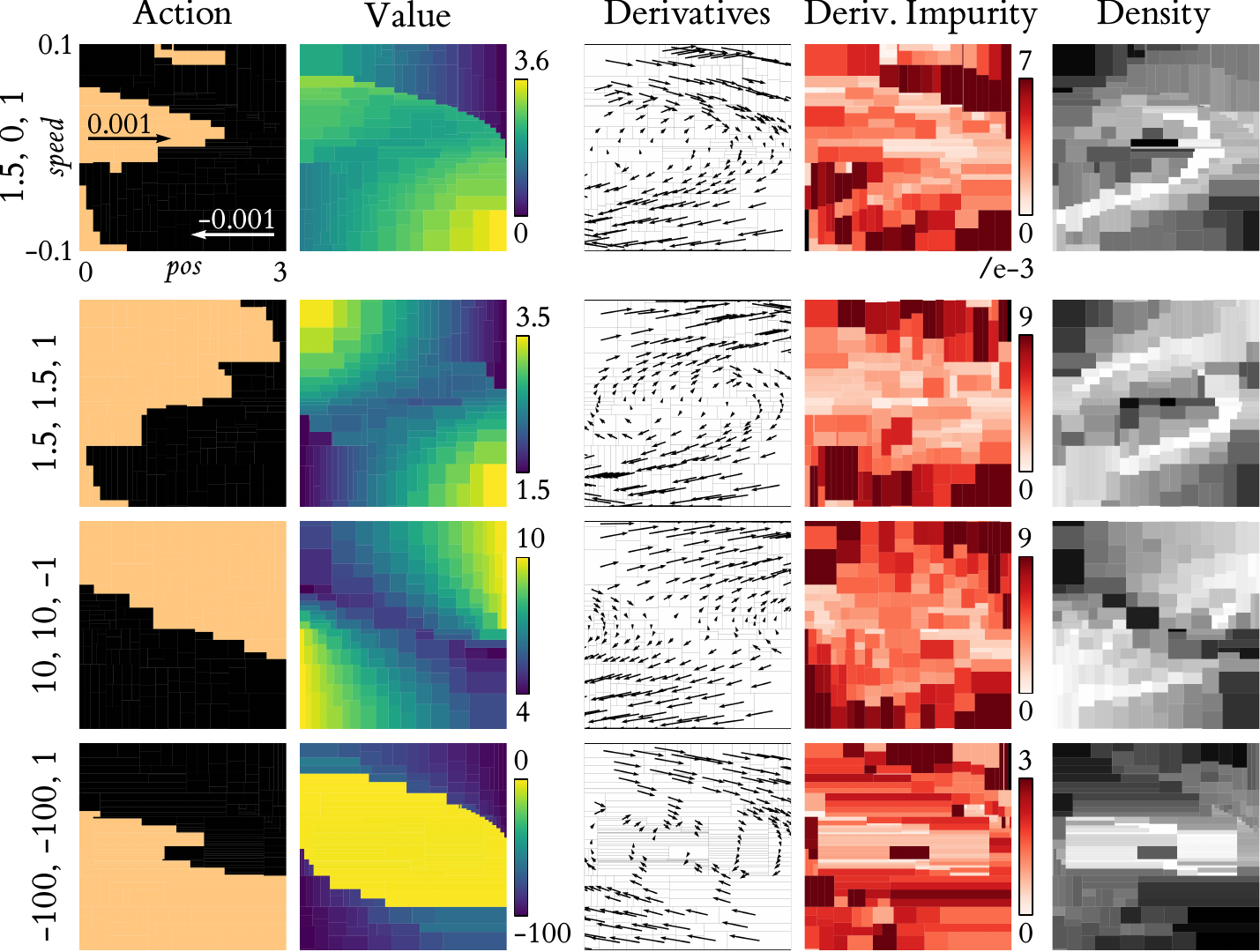}
\caption{Five types of visualisation using \textsc{TripleTree}.}
\label{fig:treemaps}
\end{figure}

Figure \ref{fig:treemaps} demonstrates the rich information conveyed by such visualisations in the road MDP. Each row of plots is generated from a single \textsc{TripleTree} with $200$ leaves, grown using the compromise weighting $\theta=[0.2,0.6,0.2]$. In the first column, leaves are coloured by predicted action, revealing the optimal DP policies. The decision boundaries have varying complexity; interesting features include the isolated `island' of positive acceleration in the top policy, which occurs when a crash with the right wall is unavoidable but positive $R_{speed}$ can be obtained by accelerating in the meantime, and the Z-shaped feature in the bottom policy, which causes the agent to oscillate around the centre of the road to avoid hitting either wall (reward $=-100$). In the second column, colours denote the predicted value, which intuitively reflects the differing reward components. In general, low value corresponds to an imminent crash into a low-reward wall. Value is high when the agent approaches a high-reward wall and/or has plenty of room to accumulate positive $R_{speed}$. For the bottom policy, value is high within a boundary of stability for the oscillatory motion, and low elsewhere. The plots of predicted derivatives in the third column differ from the others. Since this is a vector quantity, we show it as a quiver plot with an arrow for each leaf, whose direction and magnitude reflect the mean change in state between successive timesteps. The system changes more rapidly at high speeds, hence the longer arrows in these areas. Quiver plots provide an excellent high-level overview of system dynamics, particularly the locations of directional changes, cycles and regions of constancy. The fourth column colours leaves by derivative impurity, showing where in $\mathcal{S}$ we should be most confident in the model's derivative predictions. We can also create equivalent plots for action and value impurity. The final colouring attribute is sample density, computed by dividing the population of each leaf by its volume in $\mathcal{S}$: the product of boundary lengths, normalised by each feature's range across $\mathcal{D}$. This reveals where the policies spend the most time: in narrow arcs for the top two, and a tight central patch for the bottom one. 

\section{Rule-based Explanation}

\begin{figure}
\includegraphics[width=\columnwidth]{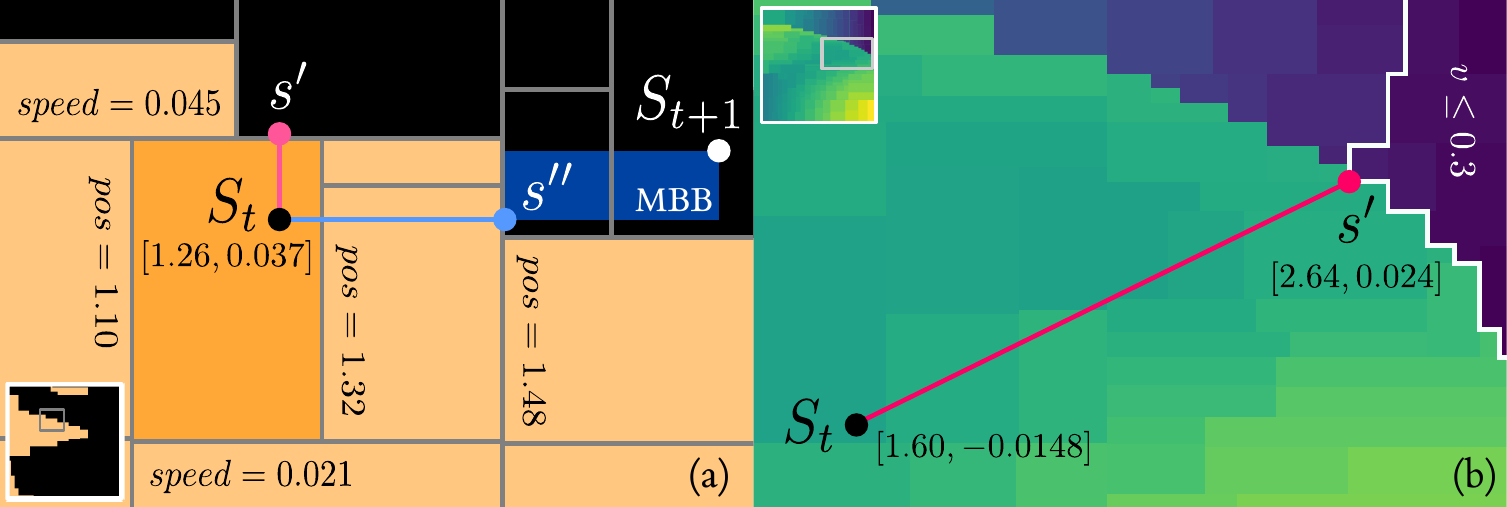}
\caption{Various forms of explanation for action and value.}
\label{fig:explanation}
\end{figure}

A popular interpretability feature of decision trees is the generation of textual explanations of outputs in terms of the decision rules applied. The simplest type of rule-based explanation is a \textit{factual} one. For any leaf $L\in\mathcal{L}$, simply enumerating the boundaries of the leaf's hyperrectangle describes the region of $\mathcal{S}$ within which a constant prediction holds. Figure \ref{fig:explanation}a shows a portion of the action visualisation for one of the $200$-leaf trees from the previous section. The action for state $S_t$ can be explained factually as follows:
\begin{center}
``Action $=0.001$ because $pos\in[1.1,1.32]$ and $speed\in[0.021,0.045]$."
\end{center}

It is argued that a more natural \cite{lipton1990contrastive} and legally persuasive \cite{wachter2017counterfactual} form of explanation is the \textit{counterfactual}, which provides reasons why an alternative outcome, known as a foil, does not occur instead. In our context, the foil is an action other than the one taken in $S_t$. After enumerating all leaves which predict the foil action, we choose one, then find the change in state required to move to a location $s'$ in that leaf. There is much debate about how to select $s'$ from many alternatives \cite{guidotti2019factual,poyiadzi2020face}, which often hinges on the notion of a \textit{minimal} change in state. In \textsc{TripleTree}, we use a two-stage process based on the $L_0$ and $L_2$ norms (Appendix B), which gives the following minimal counterfactual for the action in figure \ref{fig:explanation}a:
\begin{center}
``Action would $=-0.001$ if $speed\geq0.045$."
\end{center}

For the ordered data in $\mathcal{D}$, a third form of explanation is \textit{temporal}, which explains changes over time such as the action change from $S_t$ to $S_{t+1}$ in figure \ref{fig:explanation}a. This again takes a counterfactual perspective, although a subtlety is that using $S_{t+1}$ directly as a foil does not produce a minimal explanation. Our method for resolving this (also in Appendix B) finds $s''$, the minimal foil from $S_{t}$ subject to the constraint that the minimum bounding box (MBB) of $s''$ and $S_{t+1}$ only intersects leaves with the same action as $S_{t+1}$. This yields:
\begin{center}
``Action changed $0.001\rightarrow-0.001$ because $pos\geq1.48$."
\end{center}
Temporal explanation could be extended to a longer sequence of samples by identifying all timesteps at which the action changes, explaining each as above, and combining them into a behavioural story using the conjunction ``then". 

The \textsc{TripleTree} model allows us to similarly explain value predictions $\tilde{v}_L$. Since value is continuous, a counterfactual could explain why value is less than or greater than a threshold, rather than defining a precise numerical foil which will only ever be made by one leaf at most. In figure \ref{fig:explanation}b, we can see that the foil condition $v\leq 0.3$ leads to the following minimal counterfactual for the value at $S_t$:
\begin{center}
``Value would $\leq 0.3$ if $pos\geq2.64$ and $speed\geq0.024$."
\end{center}

\section{Trajectory Simulation}
For each leaf $L$, the derivative prediction $\tilde{d}_L$ and transition probabilities $P_L$ both describe the agent's movement through a region of $\mathcal{S}$. These can be combined to construct behavioural trajectories which may never have occurred in $\mathcal{D}$, but are nonetheless realistic given the agent's policy. This could be useful for answering targeted queries about how it navigates before, after and between states of interest. 

Here we consider the problem of finding such a trajectory between a given start leaf $L_S$ and an end leaf $L_E$. We start by using Dijkstra's algorithm \cite{dijkstra1959note} to find a sequence of leaves $\mathcal{L}_{S\rightarrow E}=(L_S,L_1,L_2,...,L_E)$ that the agent moves through with nonzero probability. We define the (inverse) cost of each transition $L\rightarrow L'$ according to the probability $P_L(L')$, and the cost of a full sequence as the product of its constituent transitions. If valid sequences exist between $L_S$ and $L_E$, Dijkstra's algorithm is guaranteed to find the highest-probability one first. If no solutions exist the algorithm returns a null result, which still gives valuable information about the non-reachability of states.  

To generate a realistic trajectory through $\mathcal{L}_{S\rightarrow E}$, we solve a constrained optimisation problem to build a piecewise linear path whose segments are well aligned with the leaves' predicted derivative vectors. Concretely, for each leaf $L_j\in\mathcal{L}_{S\rightarrow E}$ we initialise a path node $p_j$ on the hyperrectangle boundary, then perform gradient descent updates on all node locations to minimise the squared angular deviation between the path segments and the derivative vectors. When calculating angles, we normalise derivatives by the vector of inverse standard deviations across $\mathcal{D}$, $1/\sigma$. The unconstrained update to $p_j$ is proportional to the partial derivative of the squared deviations for the segments before and after:
\begin{equation}
\frac{\partial}{\partial p_j}\Big[\Big(\cos^{-1}\frac{x_j\cdot d_j}{||x_j||||d_j||}\Big)^2+\Big(\cos^{-1}\frac{x_{j+1}\cdot d_{j+1}}{||x_{j+1}||||d_{j+1}||}\Big)^2\Big],
\end{equation}
where $x_j=(p_j-p_{j-1})\circ 1/\sigma$ and $d_j=\tilde{d}_{L_j}\circ 1/\sigma$ ($\circ$ denotes the Hadamard product). Rather than applying the update directly, we constrain the node to remain on its respective boundary, and always `visible' from the previous node (i.e. it never moves to the far side of the boundary). 

\begin{figure}
\includegraphics[width=\columnwidth]{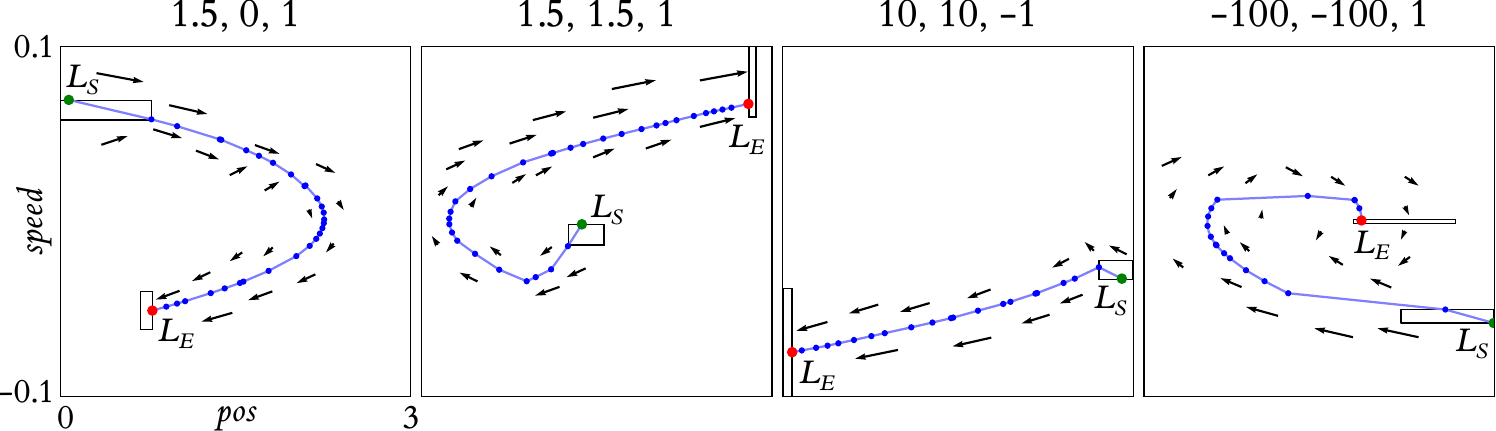}
\caption{Simulated trajectories in the road MDP.}
\label{fig:trajectories}
\end{figure}

Figure \ref{fig:trajectories} contains trajectories generated by this search-then-align method from the $200$-leaf \textsc{TripleTrees}. Derivative arrows from nearby leaves indicate that the trajectories align well with agents' true motion in each region of $\mathcal{S}$. The optimisation converges reliably and generally yields high-quality trajectories, but is rather expensive and can get stuck in local minima. We are exploring refinements to our approach, and how it may be combined with temporal explanation to provide a textual summary of simulated trajectories.

\section{Experiments in a Higher-dimensional MDP}

We now deploy \textsc{TripleTree} in a more complex MDP: \textsc{LunarLanderContinuous-v2} within OpenAI Gym \cite{Gym}. Here, the state $s$ is an $8$-dimensional vector $[x,y,v_x,v_y,\phi,v_\phi,c_L,c_R]$, which are respectively the horizontal and vertical position and velocity, orientation, and angular velocity of a landing craft, and binary flags as to whether its left and right legs contact the ground. The action space $\mathcal{A}=[-1,1]^2$ is bounded and 2D. The first component is the throttle for the lander's vertical engine ($-1$ is off) and the second is a left-right side engine ($0$ is off). The reward is $+100$ for a safe landing in a landing zone and $-100$ for a crash, and there is additional shaping reward to disincentivise fuel burn. The black box target policy for our model is a Soft Actor-Critic deep RL agent from \textit{Baselines Zoo} \cite{rl-zoo}, the highest-performing policy on that repository.  

\begin{figure}[ht]
\includegraphics[width=\columnwidth]{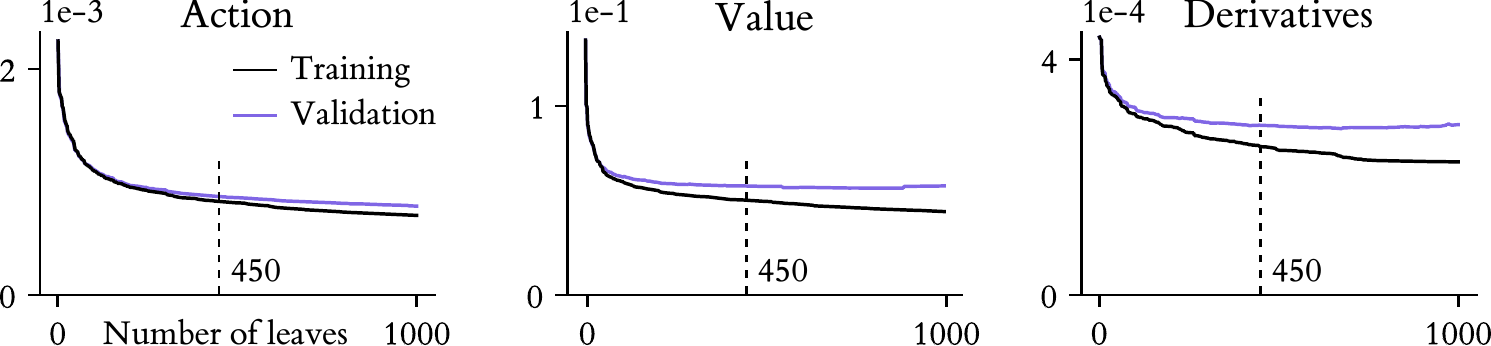}
\caption{Training and validation losses in \textsc{LunarLander}.}
\label{fig:loss_lunar}
\vspace{0.2cm}
\includegraphics[width=\columnwidth]{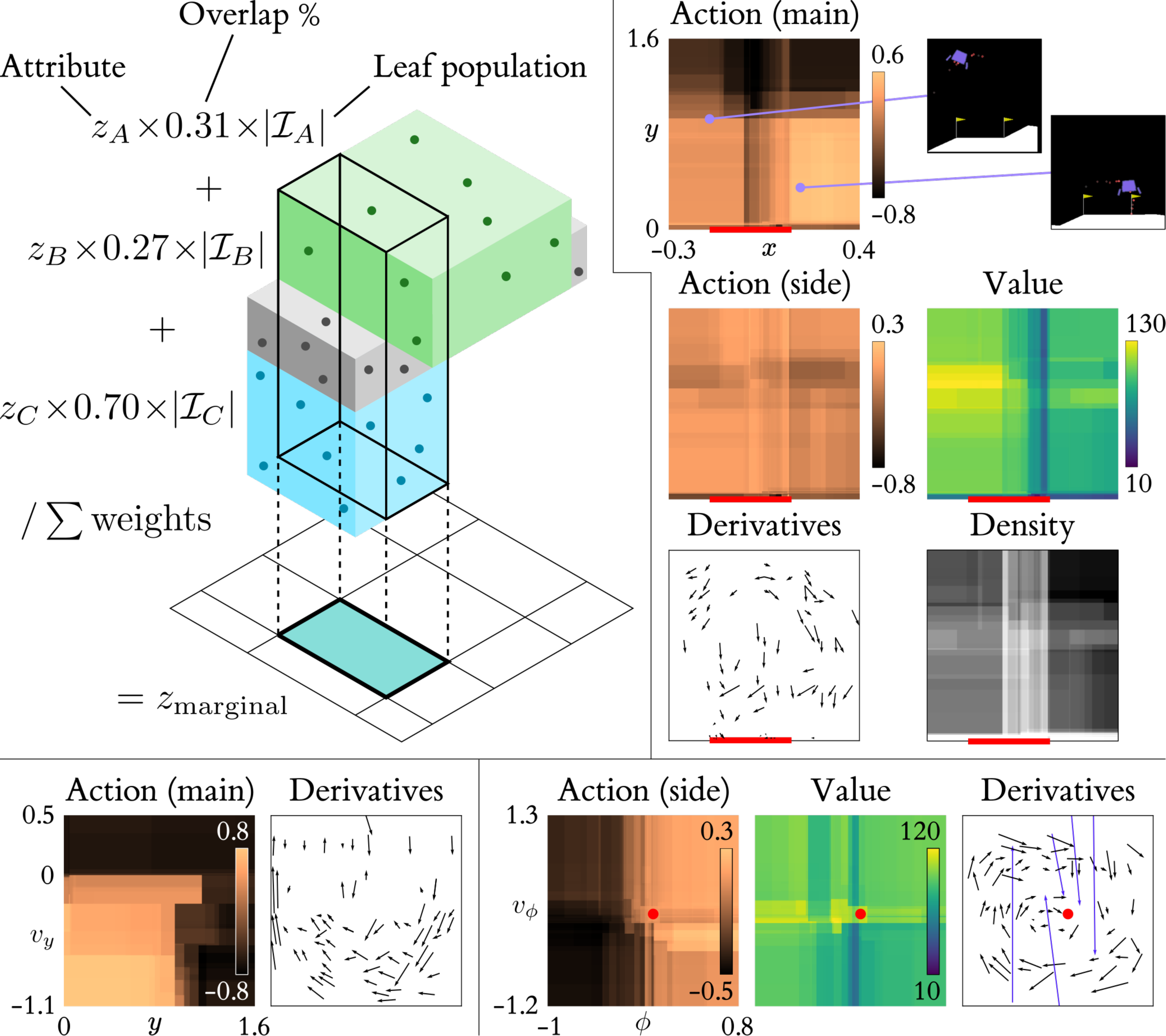}
\caption{PDP projection method and results.}
\label{fig:projection_lunar}
\vspace{0.2cm}
\includegraphics[width=\columnwidth]{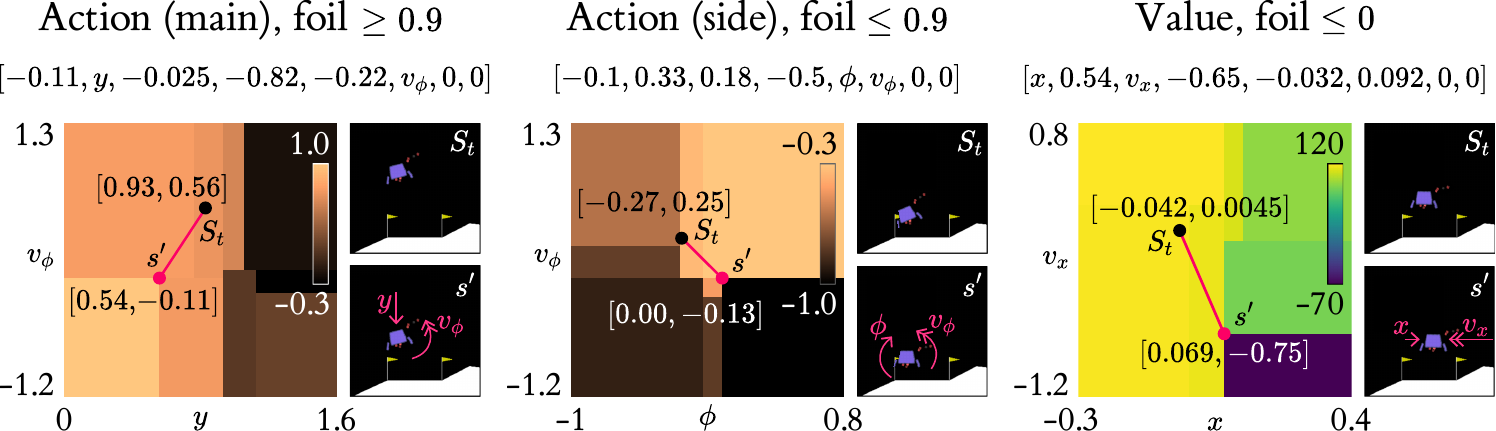}
\caption{Displaying visual counterfactuals on ICE plots.}
\label{fig:slice_and_explanation_lunar}
\vspace{0.2cm}
\includegraphics[width=\columnwidth]{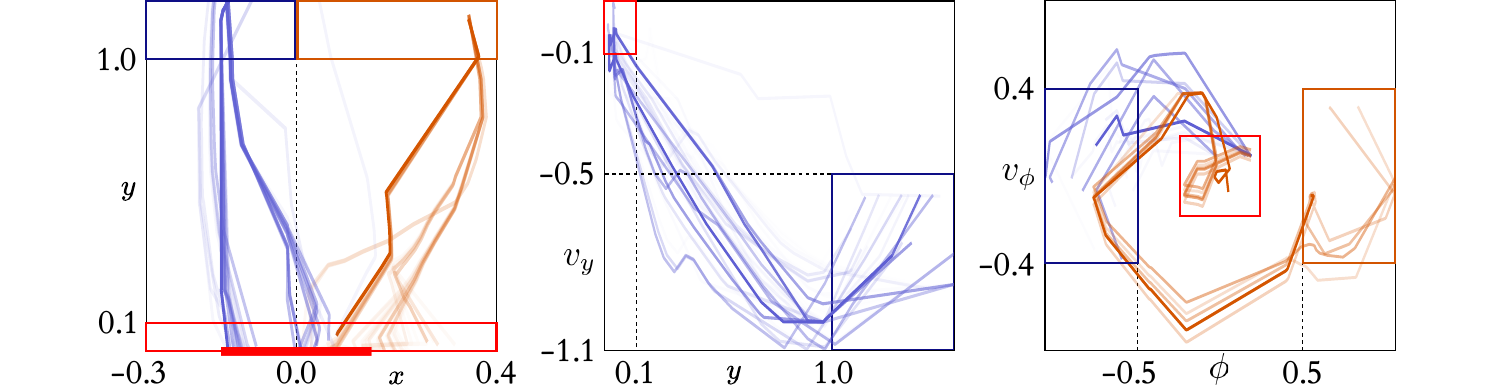}
\caption{Simulated trajectories between regions of $\mathcal{S}$. The opacity of each trajectory is proportional to its probability.
}
\label{fig:trajectories_lunar}
\end{figure}

Using a dataset of $10^5$ observations, we grow a \textsc{TripleTree} of up to $1000$ leaves with $\theta=[1,1,1]$ (the multivariate action space requires us to slightly modify the action impurity measure). Figure \ref{fig:loss_lunar} shows how the three losses vary during growth on both the training set and a validation set. In this more complex MDP the prediction problem is harder -- particularly, it seems, for derivatives -- and losses do not reduce to near zero, but as we shall see, the model still captures enough of the statistical properties of the system to deliver significant insight. We use the validation losses to inform early stopping and select the $450$-leaf tree for evaluation. 

With an 8D state space, it is nontrivial to create visualisations like those in figure \ref{fig:treemaps}. We suggest two ways forward: \textit{projection} and \textit{slicing}, which are detailed in Appendix C. In the former, we project leaf hyperrectangles onto a plane defined by two feature axes. Where multiple projections overlap, we compute a marginal value for the colouring attribute as a weighted average.  This creates a \textit{partial dependence plot} (PDP) of the attribute over the two features. Figure \ref{fig:projection_lunar} contains a diagram of the method, and results from the $450$-leaf tree. The upper five plots are PDPs for the $x$-$y$ plane (landing zone shown in red). Notice how the main engine fires less at high altitudes. Sample density is high in a column above the landing zone, and on the ground where the policy makes slow positional corrections. The value and derivatives plots reveal that despite the MDP being symmetric, the agent obtains higher value when landing from the left, and takes a less curved route when doing so. The side engine plot has weaker trends, but the dark band around $y=1$ (indicating the engine tends to fire to the left) may explain the wider landing approaches on the right. PDPs for the $y$-$v_y$ plane show hard thresholds in main engine activation at $v_y=0$ and $y\approx1$, and a U-shaped vertical speed profile. In the $\phi$-$v_\phi$ plane, we see that the side engine fires in an intuitive way to maintain stability, that value is highest when $v_\phi\approx0$, and that the lander has pendulum-like dynamics aside from several leaves (purple) where $v_\phi$ jumps. These likely reflect rapid changes in side engine activation.

Slice visualisation involves taking an axis-aligned planar cross-section of $\mathcal{S}$, and displaying all intersected leaves as rectangles. This creates an individual conditional expectation (ICE) plot of the colouring attribute rather than a PDP, which is useful for illustrating counterfactual explanations for which the true state and minimal foil differ in $\leq2$ features. Examples are shown in figure \ref{fig:slice_and_explanation_lunar}. These plots not only display the minimal state change required to realise the foil condition, but reveal some of the surrounding state space, giving an indication of the counterfactual's robustness.

We can also simulate trajectories in this MDP and visualise them in 2D. Figure \ref{fig:trajectories_lunar} contains some examples. Rather than showing a single trajectory between two leaves, we display \textit{all} possible trajectories between leaves within a
start zone (blue/orange) and end zone (red), demonstrating the distribution of paths taken by the agent. The first plot clarifies our prior observation that approaches from the right are wider, and shows that they occasionally miss the landing zone altogether. Thereafter, the lander must  `shuffle' along the ground into position; a major source of lost value. Similarly, the second plot confirms that the lander's vertical speed profile is U-shaped, and in fact very close to quadratic. The final plot is the most interesting. If rotated to the left ($\phi<0.5$), the lander's return to a stable, neutral orientation is direct and overdamped. From the right, trajectories back to neutrality tend to overshoot; a classic indicator of a poorly-tuned controller. This is further evidence that despite obtaining high reward, the policy is chronically asymmetric.

\newpage

\section{Conclusion}
G\"{a}rdenfors asserts that the aggregation of high-dimensional observational data into discrete convex regions, based similarity judgements, is a general route towards human understanding of complex systems. We consider \textsc{TripleTree} to be a practical demonstration of this phenomenon; a versatile representational tool for delivering practical insight into the behaviour of black box autonomous agents through multivariate prediction, visualisation and rule-based explanation. In ongoing work, we continue to explore the potential of this representation, refining and expanding our methods of analysis and deploying the model in more challenging MDPs.

\pagebreak
\bibliographystyle{aaai}
\bibliography{../_GLOBAL/Bibliography}

\begin{thebibliography}{30}
\providecommand{\natexlab}[1]{#1}
\providecommand{\url}[1]{\texttt{#1}}
\providecommand{\urlprefix}{URL }
\expandafter\ifx\csname urlstyle\endcsname\relax
  \providecommand{\doi}[1]{doi:\discretionary{}{}{}#1}\else
  \providecommand{\doi}{doi:\discretionary{}{}{}\begingroup
  \urlstyle{rm}\Url}\fi

\bibitem[{Adadi and Berrada(2018)}]{adadi2018peeking}
Adadi, A.; and Berrada, M. 2018.
\newblock Peeking inside the black-box: A survey on Explainable Artificial
  Intelligence (XAI).
\newblock \emph{IEEE Access} 6: 52138--52160.

\bibitem[{Bastani, Pu, and Solar-Lezama(2018)}]{bastani2018verifiable}
Bastani, O.; Pu, Y.; and Solar-Lezama, A. 2018.
\newblock Verifiable reinforcement learning via policy extraction.
\newblock In \emph{Advances in neural information processing systems},
  2494--2504.

\bibitem[{Bewley, Lawry, and Richards(2020)}]{bewley2020modelling}
Bewley, T.; Lawry, J.; and Richards, A. 2020.
\newblock Modelling Agent Policies with Interpretable Imitation Learning.
\newblock In \emph{1st TAILOR workshop at ECAI 2020}.

\bibitem[{Breiman et~al.(1984)Breiman, Friedman, Olshen, and Stone}]{CART}
Breiman, L.; Friedman, J.; Olshen, R.; and Stone, C. 1984.
\newblock {Classification and regression trees. Wadsworth \& Brooks}.
\newblock \emph{Cole Statistics/Probability Series} .

\bibitem[{Brockman et~al.(2016)Brockman, Cheung, Pettersson, Schneider,
  Schulman, Tang, and Zaremba}]{Gym}
Brockman, G.; Cheung, V.; Pettersson, L.; Schneider, J.; Schulman, J.; Tang,
  J.; and Zaremba, W. 2016.
\newblock OpenAI Gym.

\bibitem[{Carnap(1967)}]{carnap1967logical}
Carnap, R. 1967.
\newblock \emph{The logical structure of the world}.
\newblock Routledge London.

\bibitem[{Chomsky(1959)}]{chomsky1959review}
Chomsky, N. 1959.
\newblock A review of BF Skinner's Verbal Behavior.
\newblock \emph{Language} 35(1): 26--58.

\bibitem[{Coppens et~al.(2019)Coppens, Efthymiadis, Lenaerts, Now{\'e}, Miller,
  Weber, and Magazzeni}]{coppens2019distilling}
Coppens, Y.; Efthymiadis, K.; Lenaerts, T.; Now{\'e}, A.; Miller, T.; Weber,
  R.; and Magazzeni, D. 2019.
\newblock Distilling deep reinforcement learning policies in soft decision
  trees.
\newblock In \emph{Proceedings of the IJCAI 2019 Workshop on Explainable AI},
  1--6.

\bibitem[{De'Ath(2002)}]{de2002multivariate}
De'Ath, G. 2002.
\newblock Multivariate regression trees: a new technique for modeling
  species--environment relationships.
\newblock \emph{Ecology} 83(4): 1105--1117.

\bibitem[{Dijkstra(1959)}]{dijkstra1959note}
Dijkstra, E.~W. 1959.
\newblock A note on two problems in connexion with graphs.
\newblock \emph{Numerische mathematik} 1(1): 269--271.

\bibitem[{Edelman(1998)}]{edelman1998representation}
Edelman, S. 1998.
\newblock Representation is representation of similarities.
\newblock \emph{The Behavioral and brain sciences} 21(4): 449.

\bibitem[{G{\"a}rdenfors(2004)}]{gardenfors2004conceptual}
G{\"a}rdenfors, P. 2004.
\newblock \emph{Conceptual spaces: The geometry of thought}.
\newblock MIT press.

\bibitem[{Guidotti et~al.(2019)Guidotti, Monreale, Giannotti, Pedreschi,
  Ruggieri, and Turini}]{guidotti2019factual}
Guidotti, R.; Monreale, A.; Giannotti, F.; Pedreschi, D.; Ruggieri, S.; and
  Turini, F. 2019.
\newblock Factual and Counterfactual Explanations for Black Box Decision
  Making.
\newblock \emph{IEEE Intelligent Systems} .

\bibitem[{Jiang, Hwang, and Lin(2019)}]{jiang2019experience}
Jiang, W.-C.; Hwang, K.-S.; and Lin, J.-L. 2019.
\newblock An Experience Replay Method based on Tree Structure for Reinforcement
  Learning.
\newblock \emph{IEEE Transactions on Emerging Topics in Computing} .

\bibitem[{Kim et~al.(2015)Kim, Yue, Taylor, and Matthews}]{kim2015decision}
Kim, T.; Yue, Y.; Taylor, S.; and Matthews, I. 2015.
\newblock A decision tree framework for spatiotemporal sequence prediction.
\newblock In \emph{Proceedings of the 21th ACM SIGKDD International Conference
  on Knowledge Discovery and Data Mining}, 577--586.

\bibitem[{Koul, Fern, and Greydanus(2018)}]{koul2018learning}
Koul, A.; Fern, A.; and Greydanus, S. 2018.
\newblock Learning Finite State Representations of Recurrent Policy Networks.
\newblock In \emph{International Conference on Learning Representations}.

\bibitem[{Lipton(1990)}]{lipton1990contrastive}
Lipton, P. 1990.
\newblock Contrastive explanation.
\newblock \emph{Royal Institute of Philosophy Supplements} 27: 247--266.

\bibitem[{Liu et~al.(2018)Liu, Schulte, Zhu, and Li}]{liu2018toward}
Liu, G.; Schulte, O.; Zhu, W.; and Li, Q. 2018.
\newblock Toward interpretable deep reinforcement learning with linear model
  u-trees.
\newblock In \emph{Joint European Conference on Machine Learning and Knowledge
  Discovery in Databases}, 414--429. Springer.

\bibitem[{Poyiadzi et~al.(2020)Poyiadzi, Sokol, Santos-Rodriguez, De~Bie, and
  Flach}]{poyiadzi2020face}
Poyiadzi, R.; Sokol, K.; Santos-Rodriguez, R.; De~Bie, T.; and Flach, P. 2020.
\newblock FACE: feasible and actionable counterfactual explanations.
\newblock In \emph{Proceedings of the AAAI/ACM Conference on AI, Ethics, and
  Society}, 344--350.

\bibitem[{Pyeatt(2003)}]{pyeatt2003reinforcement}
Pyeatt, L.~D. 2003.
\newblock Reinforcement learning with decision trees.
\newblock In \emph{21st IASTED International Multi-Conference on Applied
  Informatics}, 26--31.

\bibitem[{Raffin(2018)}]{rl-zoo}
Raffin, A. 2018.
\newblock RL Baselines Zoo.
\newblock \url{https://github.com/araffin/rl-baselines-zoo}.

\bibitem[{Rosch et~al.(1976)Rosch, Mervis, Gray, Johnson, and
  Boyes-Braem}]{rosch1976basic}
Rosch, E.; Mervis, C.~B.; Gray, W.~D.; Johnson, D.~M.; and Boyes-Braem, P.
  1976.
\newblock Basic objects in natural categories.
\newblock \emph{Cognitive psychology} 8(3): 382--439.

\bibitem[{Roth et~al.(2019)Roth, Topin, Jamshidi, and
  Veloso}]{roth2019conservative}
Roth, A.~M.; Topin, N.; Jamshidi, P.; and Veloso, M. 2019.
\newblock Conservative Q-Improvement: Reinforcement Learning for an
  Interpretable Decision-Tree Policy.
\newblock \emph{arXiv:1907.01180} .

\bibitem[{Saghezchi and Asadpour(2010)}]{saghezchi2010multivariate}
Saghezchi, H.~B.; and Asadpour, M. 2010.
\newblock Multivariate decision tree function approximation for reinforcement
  learning.
\newblock In \emph{International Conference on Neural Information Processing},
  687--694. Springer.

\bibitem[{Samek, Wiegand, and M{\"u}ller(2017)}]{samek2017explainable}
Samek, W.; Wiegand, T.; and M{\"u}ller, K.-R. 2017.
\newblock {Explainable AI: Understanding, visualizing and interpreting deep
  learning models}.
\newblock \emph{arXiv:1708.08296} .

\bibitem[{Sutton and Barto(2018)}]{sutton2018reinforcement}
Sutton, R.~S.; and Barto, A.~G. 2018.
\newblock \emph{Reinforcement learning: An introduction}.
\newblock MIT press.

\bibitem[{Thomas and Okal(2015)}]{thomas2015notation}
Thomas, P.~S.; and Okal, B. 2015.
\newblock A notation for Markov decision processes.
\newblock \emph{arXiv:1512.09075} .

\bibitem[{Uther and Veloso(1998)}]{uther1998tree}
Uther, W.~T.; and Veloso, M.~M. 1998.
\newblock Tree based discretization for continuous state space reinforcement
  learning.
\newblock In \emph{{AAAI/IAAI}}, 769--774.

\bibitem[{Wachter, Mittelstadt, and Russell(2017)}]{wachter2017counterfactual}
Wachter, S.; Mittelstadt, B.; and Russell, C. 2017.
\newblock Counterfactual explanations without opening the black box: Automated
  decisions and the GDPR.
\newblock \emph{Harv. JL \& Tech.} 31: 841.

\bibitem[{Wan et~al.(2020)Wan, Dunlap, Ho, Yin, Lee, Jin, Petryk, Bargal, and
  Gonzalez}]{wan2020nbdt}
Wan, A.; Dunlap, L.; Ho, D.; Yin, J.; Lee, S.; Jin, H.; Petryk, S.; Bargal,
  S.~A.; and Gonzalez, J.~E. 2020.
\newblock NBDT: Neural-Backed Decision Trees.
\newblock \emph{arXiv:2004.00221} .

\end{thebibliography}



\section*{Supplementary material}

## Appendix A: Worst-Case Loss Analysis

To more deeply understand the tradeoff between the three types of loss in the 2D road MDP, we consider 21 equally-spaced weighting vectors $\theta$. For each, we calculate which loss is *worst* as a ratio of the loss from a tree with just one leaf, which predicts the dataset average. The results are plotted in barycentric coordinates (simplex plots) in figure 1. The prevalence of red dots throughout the range of tree sizes shows that value prediction is weakest across most of the space of weightings. A notable exception is the right-hand edge of the simplex, where derivative weight is 0 and that loss is accordingly the worst. The greyscale heatmaps show the magnitude of the worst loss ratio (interpolated using Matplotlib's `LinearTriInterpolator`), which as a general rule is lower towards the centre of the simplex, but also towards the right-hand side, where the value weight is higher. These results point to the conclusion that in this MDP, value should be up-weighted in the hybrid quality metric if minimising the worst-case loss ratio is important. From looking at the various heatmaps, we suggest that $\theta = [0.2, 0.6, 0.2]$ is close to the best possible compromise. We use this weighting throughout the rest of the main paper, aside from for the final section on LUNARLANDER.

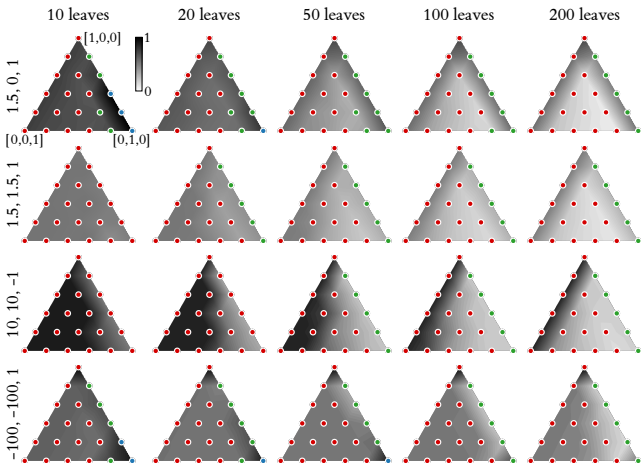

Figure 1: Analysis of worst-case loss across the space of weight vectors for trees of various sizes. Coloured dots show the identity of the worst loss (blue: action, red: value, green: derivatives) at the 21 weightings tested. Heatmaps show the magnitude of the worst loss as a ratio of the loss from a one-leaf tree, linearly interpolated between test locations.

## Appendix B: Algorithms for Explanation

### Finding the Minimal Counterfactual Foil

Let $L^*$ be the leaf where the observation to be explained, $S_t$, ends up after being propagated through the tree, and let $\tilde{z}_{L^*}$ generically represent the leaf's predicted attribute. In a TRIPLETREE this attribute may be an action, value or derivative vector. For counterfactual explanation, we specify a foil condition $F$, which is used to identify a set of leaves in which the foil state may reside. If the attribute is a discrete action, the foil simply specifies another member of the action space. For continuous action or value explanation, the foil is an inequality which is not satisfied by $\tilde{z}_{L^*}$[1]. In either case, let $\mathcal{L}' = \{L \in \mathcal{L} : F(L) = \text{True}\}$ be the subset of leaves satisfying $F$.

For each $L \in \mathcal{L}'$, let $l_L^{(f)}$ and $u_L^{(f)}$ be the lower and upper boundaries of that leaf's hyperrectangle along feature $f$, which correspond to partitions made at its ancestor nodes. One or both of these boundaries will be undefined (effectively infinite) if none of the ancestors partition along $f$. In such cases we replace a lower bound with the minimum value of the feature across all samples in $\mathcal{D}$, denoted by $s_{\min}^{(f)}$, and replace an upper bound with the maximum value $s_{\max}^{(f)}$.

The location in $L$ which is closest[2] to the input state $S_t$, denoted by $s_L$, lies on the boundary. It is defined on a feature-wise basis as follows:

$$s_L^{(f)} = \begin{cases} u_L^{(f)} & \text{if } S_t^{(f)} > u_L^{(f)}, \\ l_L^{(f)} & \text{if } S_t^{(f)} < l_L^{(f)}, \\ S_t^{(f)} & \text{otherwise.} \end{cases} \quad (1)$$

Let $\delta_L = (s_L - S_t) \cdot \alpha$ be the vector from $S_t$ to $s_L$, normalised by $\alpha$, which is the vector of the reciprocal min-max ranges of the state features across the training set $\mathcal{D}$. As discussed in the main paper, such normalisation is important to bring the state features onto equivalent scales.

The task of finding a minimal foil consists in selecting one element of the set $\{s_L : L \in \mathcal{L}'\}$ by consideration of the corresponding $\delta$ vectors. Since this set is non-convex in general, there is not one choice which is unambiguously closest to $S_t$; this depends on which norm we apply to the $\delta$s. Different norms have different advantages. The 2-norm is the most common and intuitive distance metric in vector spaces, especially those with a physical interpretation, but does not incentivise sparsity, which would allow us to give a more compact explanation (fewer feature changes means fewer clauses in the textual summary). Conversely, the 0-norm only measures sparsity, and cannot differentiate between multiple options with the same number of nonzero elements. The 1-norm offers a popular compromise, but we do not use this in TRIPLETREE. Instead, we filter first by 0-norm

$$\mathbb{S} = \{s_L : ||\delta_L||_0 = \inf\{||\delta_{L'}||_0 : L' \in \mathcal{L}'\}\}, \quad (2)$$

then by 2-norm to find the minimal foil

$$s' = s_L \in \mathbb{S} : ||\delta_L||_2 = \inf\{||\delta_{L'}||_2 : \delta_{L'} \in \mathcal{L}'\}. \quad (3)$$

This two-stage approach enforces a strict priority of the 0-norm over the 2-norm; a foil state is only considered if its $\delta$ vector is at least as sparse as any of the others. We suggest that this is desirable behaviour, because it puts the strongest possible emphasis on compact explanations, while still punishing $\delta$ vectors with very large (Euclidean) magnitudes. It

---

[1] For counterfactual explanation of derivative vectors we would need to specify an inequality for one or more features individually. We do not consider this case here.

[2] Because hyperrectangles are convex, $s_L$ is unambiguously the closest point to $S_t$ as measured by any $p$-norm.

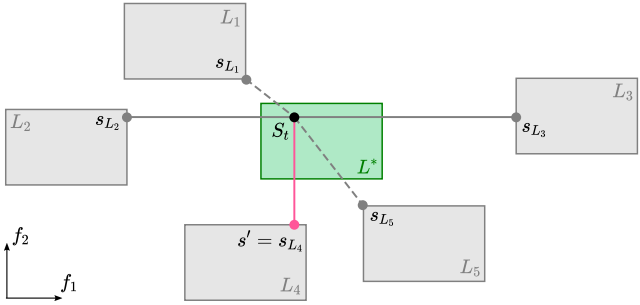

Figure 2: To find the minimal counterfactual for $S_t$, we first identify the set of leaves satisfying the foil condition $\mathcal{L}' = \{L_1, ..., L_5\}$, then find the closest point $S_t$ on the boundary of each. We first filter by the 0-norm of the corresponding $\delta$ vectors; this removes $s_{L_1}$ and $s_{L_5}$ from contention. We then filter by 2-norm, which identifies $s_{L_4}$ as the minimal foil. Assume that in this diagram, the two feature axes $f_1$ and $f_2$ have been scaled according to the normalisation vector $\alpha$.

also makes it simple, if required, to specify a hard threshold on sparsity, allowing us to answer questions of the form "can the foil condition be realised by changing $\leq n$ features?". The method is summarised diagrammatically in figure 2, for which the generic textual explanation is

"$F$ would be satisfied if $f_2 \leq s_{L_4}^{(f_2)}$."

**Adaptation for Temporal Explanation**

In a temporal explanation scenario we are given two consecutive observations $S_t$ and $S_{t+1}$ which fall into different leaves, so that $S_{t+1}$ satisfies some foil condition $F$ (e.g. different discrete action, inequality of value) with respect to $S_t$. Simply using $S_{t+1}$ itself as the foil is uninformative and tautological, and we should again consider the notion of a minimal foil. However, merely following the method described above does not solve the problem, as illustrated in figure 3.

Here $\mathcal{L}' = \{L_1, ..., L_6\}$, but every one of $s_{L_1}, ..., s_{L_6}$ is a suboptimal foil. In regular counterfactual explanation, $s_{L_1}$ would be minimal, but it implies moving in the *opposite direction* from $S_{t+1}$ so does not lie on any plausible path between the two observations. $s_{L_2}$ and $s_{L_3}$ are at least closer to $S_{t+1}$ than $S_t$ is, but in both cases there is not an unbroken path of foil leaves connecting them to $S_{t+1}$, so they do not represent *sufficient* conditions for the change in attribute. More subtly, the same is true for $s_{L_4}$; the orange shading highlights a region of $\mathcal{S}$ below and to the right of $s_{L_4}$, but which is not part of any $L \in \mathcal{L}'$. The existence of this region means that a counterfactual of the form

"$F$ became satisfied because $f_1 \geq s_{L_4}^{(f_1)}$ and $f_2 \leq s_{L_4}^{(f_2)}$."

would be *misleading*, since it is possible to satisfy the stated conditions in a state between $S_t$ and $S_{t+1}$ while not satisfying $F$. In contrast, both are $s_{L_5}$ and $s_{L_6}$ are acceptable according to this unbroken path criterion, and since the former is closer to $S_t$ according to the 2-norm, it may initially appear that this is the best possible foil.

However, we can do better than this. We propose that the best foil to use for temporal explanation, $s''$, is the location

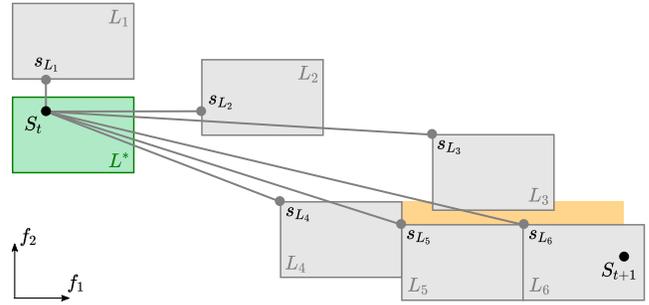

Figure 3: For temporal explanation, the vanilla counterfactual method is insufficient. All of $s_{L_1}, ..., s_{L_6}$ yield either irrelevant, misleading or non-minimal explanations of a prediction change between $S_t$ and $S_{t+1}$.

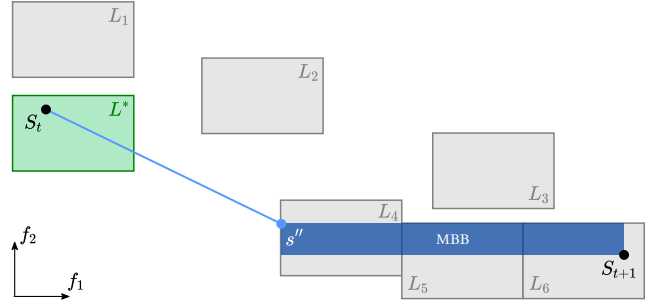

Figure 4: $s''$ is the minimal foil state from $S_t$, subject to the MBB constraint from $S_{t+1}$. We claim this this is the optimal foil to use for temporal explanation.

inside one of the foil leaves which is minimal from $S_t$, subject to the constraint that the *minimum axis-aligned bounding box* (MBB) of $s''$ and $S_{t+1}$ only intersects leaves in $\mathcal{L}'$. This constraint ensures that the generated counterfactual explanation is never misleading in the sense outlined above. Figure 4 demonstrates the best foil state in our example, for which the textual explanation is

"$F$ became satisfied because $f_1 \geq s''^{(f_1)}$ and $f_2 \leq s''^{(f_2)}$."

Finding $s''$ in practice is not trivial, and we have currently only implemented a method for 2D state spaces, outlined in algorithm 1. This algorithm returns $\mathcal{R}$, a set of rectangles tiling the region of $\mathcal{S}$ between $S_t$ and $S_{t+1}$ which satisfies the MBB constraint. We then apply the equations of the preceding section[3] to $\mathcal{R}$ instead of the underlying foil leaves $\mathcal{L}'$, thereby obtaining the desired foil state. In the simple example in figure 4, $\mathcal{R}$ has just a single element – the rectangle shown in blue – but in general the region satisfying the MBB constraint may be rather more complex.

## Appendix C: Visualisation for High-dimensional MDPs

**Projection**

In the projection visualisation method, we choose two feature axes and project leaf hyperrectangles onto the plane de-

---

[3]The placeholder function "`rectangle`" in algorithm 1 creates a representation of each rectangle which can be fed into these equations in the same way as the leaves themselves.

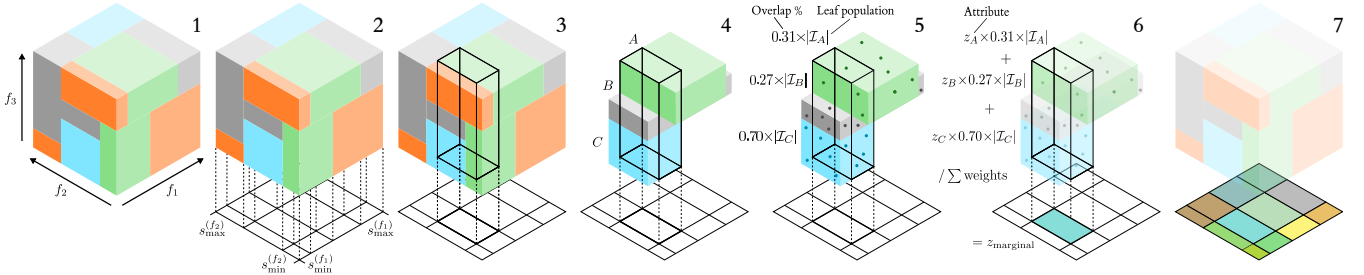

Figure 5: Steps of the hyperrectangle projection process, with $d = 3$ features for ease of presentation. Colours represent the leaf-level attribute to be visualised, such as predicted action or value. $z$ is a generic placeholder to the attribute to be visualised.

fined by these axes. By taking a population-weighted average of the attributes from overlapping leaf projections, we effectively create a partial dependence plot (PDP).

Let $\mathcal{B}^{(f)}$ denote the complete sequence of hyperrectangle boundaries in the tree along feature $f$, sorted by value:

$$\mathcal{B}^{(f)} = \left( b_i \in \bigcup_{L \in \mathcal{L}} \{l_L^{(f)}, u_L^{(f)}\} \right),$$

$$\text{subject to } \forall i \in \{2..|\mathcal{B}^{(f)}|\} \; \mathcal{B}_i^{(f)} > \mathcal{B}_{i-1}^{(f)}. \quad (4)$$

where $l_L^{(f)}$ and $u_L^{(f)}$ are defined as above, with any undefined boundaries again replaced by $s_{\min}^{(f)}$ or $s_{\max}^{(f)}$.

Assume without loss of generality that the two features onto which we are projecting are $f_1$ and $f_2$. The planar region $s_{\min}^{(f_1)} \leq s^{(f_1)} \leq s_{\max}^{(f_1)}$, $s_{\min}^{(f_2)} \leq s^{(f_2)} \leq s_{\max}^{(f_2)}$ can be tiled by a $(|\mathcal{B}^{(f_1)}| - 1) \times (|\mathcal{B}^{(f_2)}| - 1)$ grid of rectangles, each of which is constructed from two pairs of consecutive boundaries from $\mathcal{B}^{(f_1)}$ and $\mathcal{B}^{(f_2)}$. Steps 1 and 2 in figure 5 illustrate this reasoning.

If each rectangular area were to be simultaneously extruded along each of the $d - 2$ orthogonal feature axes, the volume swept (which we call a *core*) would itself be a $d$-dimensional hyperrectangle which intersects with at least one of the leaf hyperrectangles. For each $i \in \{2..|\mathcal{B}^{(f_1)}|\}$, $j \in \{2..|\mathcal{B}^{(f_2)}|\}$, the set of intersected hyperrectangles can be identified as

$$\mathcal{L}_{i,j} = \Big\{ L \in \mathcal{L} :$$
$$(l_L^{(f_1)} \leq \mathcal{B}_{i-1}^{(f_1)}) \wedge (u_L^{(f_1)} \geq \mathcal{B}_i^{(f_1)}) \wedge$$
$$(l_L^{(f_2)} \leq \mathcal{B}_{j-1}^{(f_2)}) \wedge (u_L^{(f_2)} \geq \mathcal{B}_j^{(f_2)}) \Big\}. \quad (5)$$

Steps 3 and 4 in figure 5 show the process of extruding a rectangle from the $f_1$-$f_2$ plane and identifying intersections. In this case the core intersects the three leaves labelled A, B and C.

Precisely how we proceed from this point depends which attribute we wish to visualise. For the sake of brevity, we assume we are visualising predicted actions $\tilde{a}$ and that the action space is continuous[4]. For each rectangle in the $f_1$-$f_2$ plane, identified by boundary indices $i$ and $j$ as above, we

[4]The process is identical for visualising value predictions and impurities, and for derivative predictions we perform the aver-

effectively marginalise out the $d - 2$ orthogonal dimensions by taking a weighted mean of the predicted actions from the intersected leaves $\mathcal{L}_{i,j}$. The weight for each leaf $L \in \mathcal{L}_{i,j}$, is jointly determined by its *population* $|\mathcal{I}_L|$ and the degree to which its hyperrectangle *overlaps* with the core. We can therefore define the projected action prediction for rectangle $i, j$ as

$$\tilde{a}_{i,j} = \frac{\sum_{L \in \mathcal{L}_{i,j}} w_L^{(i,j)} \tilde{a}_L}{\sum_{L \in \mathcal{L}_{i,j}} w_L^{(i,j)}} \quad (6)$$

where

$$w_L^{(i,j)} = |\mathcal{I}_L| \left[ \frac{\mathcal{B}_i^{(f_1)} - \mathcal{B}_{i-1}^{(f_1)}}{u_L^{(f_1)} - l_L^{(f_1)}} \right] \left[ \frac{\mathcal{B}_j^{(f_2)} - \mathcal{B}_{j-1}^{(f_2)}}{u_L^{(f_1)} - l_L^{(f_1)}} \right]. \quad (7)$$

Due to the way in which the underlying tiling of the $f_1$-$f_2$ plane is defined, the core must contain exactly zero boundaries along either $f_1$ or $f_2$, so both fractions in the formula for $w_L^{(i,j)}$ are always $\leq 1$. This part of the process is illustrated by steps 5 and 6 of figure 5, and step 7 shows the result of repeating for all remaining rectangles on the plane, thereby creating the 2D visualisation.

The key assumption behind this core-and-average approach to projection is that samples are close to uniformly distributed within leaf hyperrectangles, so that each $w_L^{(i,j)}$ is an unbiased estimate of the number of samples from $L$ within the core, and each $\tilde{a}_L$ is an unbiased estimate of the mean action for those samples.

**Slicing**

In the slicing visualisation method we again choose two features to visualise over, $f_1$ and $f_2$. For each remaining feature $f_i$ (where $i \in \{3..d\}$), we specify a single threshold between $s_{\min}^{(f_i)}$ and $s_{\max}^{(f_i)}$, denoted by $s_{\text{slice}}^{(f_i)}$. The set of thresholds defines an axis-aligned planar cross-section through $\mathcal{S}$, which

aging on an elementwise basis. For discrete actions, we cannot take a weighted mean so instead add up the per-action counts $\text{count}(\mathcal{I}_L, a)$ for overlapping leaves, weighted by their overlap proportions, then visualise the modal action for each rectangle. For density, we replace the population factor $|\mathcal{I}_L|$ in equation 7 with the leaf's feature-scaled volume in $\mathcal{S}$, as defined in the main paper. We also find that it is best to use a logarithmic colour map for density plots, since this attribute can vary over many orders of magnitude between leaves.

intersects a subset of the leaves. Here we do not have to handle overlaps, and can display the rectangular cross-sections of the intersected leaves directly. This creates an individual conditional expectation (ICE) plot.

The slicing process is straightforward. Given the thresholds $\mu^{(f_3)}, ..., \mu^{(f_d)}$, we simply need to identify the subset of intersected leaves:

$$\mathcal{L}_{\text{slice}} = \bigcap_{i=3}^{d} \left\{ L \in \mathcal{L} : (l_L^{(f_i)} \leq s_{\text{slice}}^{(f_i)} \wedge (u_L^{(f_1)} \geq s_{\text{slice}}^{(f_i)}) \right\}. \tag{8}$$

We then visualise each $L \in \mathcal{L}_{\text{slice}}$ as a rectangle with boundaries at $l_L^{(f_1)}$, $u_L^{(f_1)}$, $l_L^{(f_2)}$ and $u_L^{(f_2)}$, and coloured according to its attribute.

## A Note on Unifying Projection and Slicing

In the preceding discussion, we have presented projection and slicing as two distinct visualisation methods, but in reality it is possible to smoothly transition between the two. Starting from the projection method as described, we achieve this by permitting the core extrusion along feature $f_i$ to be limited between two thresholds $s_{\text{low}}^{(f_i)} \geq s_{\text{min}}^{(f_i)}$ and $s_{\text{high}}^{(f_i)} \leq s_{\text{max}}^{(f_i)}$. This allows use to visualise projections from only those leaves within a hyperrectangular subset of $\mathcal{S}$ rather than the entire space. From this point, it is easy to see that slicing results from the limiting case where

$$s_{\text{low}}^{(f_i)} = s_{\text{high}}^{(f_i)} = s_{\text{slice}}^{(f_i)} \quad \forall i \in \{3..d\}. \tag{9}$$

---

**Algorithm 1:** Finding the region of $\mathcal{S}$ satisfying the MBB constraint (for $d = 2$ only)

**Input:** Observations $S_t$ and $S_{t+1}$; the sets of all leaves $\mathcal{L}$ and foil leaves $\mathcal{L}'$.
**Result:** A set of rectangles $\mathcal{R}$.

Initialise $s_1 \leftarrow S_{t+1}, \mathcal{R} \leftarrow \{\}$
**while** True **do**
$\quad s_2 \leftarrow s_1$
$\quad$ // Expand rectangle along both feature axes.
$\quad s_2^{(1)} \leftarrow \text{extend}(1, 2, s_1, s_2)$
$\quad s_2^{(2)} \leftarrow \text{extend}(2, 1, s_1, s_2)$
$\quad$ // Break if no expansion.
$\quad$ **if** $s_1^{(2)} = s_2^{(2)}$ **break**
$\quad$ // Store rectangle and move $s_1$ along axis 2 to reset
$\quad \mathcal{R} \leftarrow \mathcal{R} \cup \{\text{rectangle}(s_1, s_2)\}$
$\quad s_1^{(2)} \leftarrow s_2^{(2)}$
**end**
**return** $\mathcal{R}$

**Function** extend(*a*, *b*, $s_1$, $s_2$):
$\quad$ /* Given a rectangle with opposite corners at $(s_1, s_2)$, move $s_2$ along axis $a$ until the rectangle either intersects a non-foil leaf $\in \mathcal{L}_{nf}$ or extends beyond $S_t^{(a)}$. */
$\quad l, u \leftarrow \inf\{s_1^{(b)}, s_2^{(b)}\}, \sup\{s_1^{(b)}, s_2^{(b)}\}$
$\quad \mathcal{L}_{nf} \leftarrow \{L \in \mathcal{L} \setminus \mathcal{L}' : l_L^{(b)} \leq u \wedge u_L^{(b)} \geq l\}$
$\quad$ **if** $S_t^{(a)} > S_{t+1}^{(a)}$ **then**
$\quad\quad$ // Extend in the positive direction.
$\quad\quad$ **return**
$\quad\quad\quad \inf\{l_L^{(a)} : L \in \mathcal{L}_{nf} \wedge l_L^{(a)} \geq s_1^{(a)}\} \cup \{S_t^{(a)}\}$
$\quad$ **else**
$\quad\quad$ // Extend in the negative direction.
$\quad\quad$ **return**
$\quad\quad\quad \sup\{u_L^{(a)} : L \in \mathcal{L}_{nf} \wedge u_L^{(a)} \leq s_1^{(a)}\} \cup \{S_t^{(a)}\}$
$\quad$ **end**



\end{document}